\documentclass{article}

\usepackage[utf8]{inputenc}
\usepackage{cite}
\usepackage{amsmath,amssymb,amsfonts, mathtools}
\usepackage{algorithmic}
\usepackage{graphicx}
\usepackage{textcomp}
\usepackage{xcolor}
\usepackage{siunitx}
\usepackage{url}
\usepackage{subcaption}
\usepackage{chngpage}

\graphicspath{ {./static/} } 

\title{Use of Remote Sensing Data to Identify Air Pollution Signatures in India}
\author{Sivaramakrishnan KN\footnote{f20170153@hyderabad.bits-pilani.ac.in, BITS Pilani Hyderabad Campus, India}  , Lipika Deka \footnote{lipika.deka@dmu.ac.uk, Senior Lecturer, De Montfort University, United Kingdom, IEEE/ACM Professional Member} , Manik Gupta \footnote{manik@hyderabad.bits-pilani.ac.in, Assistant Professor, BITS Pilani Hyderabad Campus, India, IEEE/ACM Professional Member, APPCAIR member} }

\begin{document}

\maketitle

\begin{abstract}
Air quality has major impact on a country's socio-economic position and identifying major air pollution sources is at the heart of tackling the issue. Spatially and temporally distributed air quality data acquisition across a country as varied as India has been a challenge to such analysis. The launch of the Sentinel-5P satellite has helped in the observation of a wider variety of air pollutants than measured before at a global scale on a daily basis. In this chapter, spatio-temporal multi pollutant data retrieved from Sentinel-5P satellite is used to  cluster states as well as districts in India and  associated average monthly pollution signature and trends depicted by each of the clusters are derived and presented.The clustering signatures can be used to identify states and districts based on the types of pollutants emitted by various pollution sources.

\end{abstract}

\section{Introduction}

Air pollution is one of the major health hazards in a developing country such as India. Hence, it is necessary to study the composition of the air over the districts of our country to understand how to tackle pollutants at an individual level. Zheng et.al \cite{zheng} has shown that the concentration of particulate matter less than 2.5 micrometers in diameter ($PM2.5$) has a direct positive correlation to the number of cases of lung diseases such as asthma in patients. Other pollutants which directly affect human respiratory system are nitrogen dioxide, sulphur dioxide, formaldehyde and tropospheric ozone. High concentrations of carbon monoxide can cause dizziness, confusion, severe brain damage or even death. Methane is another important gas which contributes to the greenhouse effect and it is 80 times more harmful than $CO_2$ if it sustains in the air for long periods of time. Hence, there is a dire need to equip policy makers and policy enforcers with data driven knowledge of major pollutant sources and their geo-location, as well as increase the awareness about the ill-effects of these pollutants among the general public to ensure the well being of the society.\par 
Air-pollutant sources include road-traffic, industries such as brick-kilns, portable power generators such as diesel power generators etc. \cite{cprindia}. The information presented in \cite{urbanemissions}  details currently used methods for identifying pollution sources. Methods can be expensive and may fail to detect all sources as they may be located in geographical wide areas. Recent efforts to identify pollution sources involved the use of low-cost sensors \cite{mitnews}. However, such methods need the additional overhead of sensor installation and maintenance of each site.\par 

The aim of this chapter is to present clustering methods on air pollution data derived from larger granularity satellite data and to identify the sources behind the pollution in each cluster. Grouping of the various states and districts in India has been done based on the pollution signatures emitted from different regions. It is cost effective to use satellite data since it does not involve additional installation overheads and provides a more holistic wider area picture. Such an approach can also be used to  provide an insight for site selection for low cost sensor installation for more accurate and localised studies if required.\par

The outcome will aid the government in reforming policies to help alleviate the pollution levels effectively. Once it is determined which sources are responsible for emitting a particular pollutant, government bodies can take necessary steps to reduce the specific pollutant at the specific geo-location by placing restrictions targeting only the relevant sources. This will help the government to efficiently formulate targeted and more effective strategies to combat individual pollutants levels if there is an alarming spike in the levels of a specific pollutant.

Rest of the chapter is arranged as follows: section \ref{lit review} touches upon the most notable work done in the area till date; section \ref{collection} details the method employed to collect the required data used in this study, with section \ref{preprocessing} explaining how the data was prepared and preprocessed for further analysis; section \ref{clustering} presents the clustering algorithms employed with the discussion on the results achieved and finally conclusions in section \ref{conclusion} mentioning briefly now how the research can be taken ahead. 

\section{Literature Review} \label{lit review}

Air pollution monitoring has been a major challenge faced by the Government of India, especially considering the fact that 26 of the 50 most polluted cities in the world are located in India as of 2019\cite{world cities}. $PM2.5$, particulate matter of aerodynamic diameter less than 2.5 $\mu$m, was found to cause damage to the air passages, cardiovascular impairments, increase the likelihood of Diabetes Mellitus in humans and cause adverse effects in infancy \cite{feng}. Thus, with the advent of low cost sensors \cite{low cost sensor}, it has become possible to evaluate the air quality at a given location by only taking into consideration the $PM2.5$ levels. But there are other pollutants as well that can cause adverse health effects, such as $NO_2$, $SO_2$ and $CO$, which can be measured by a high quality air pollution monitoring system, but which requires a higher installation cost and maintenance.

As of 2020, there are only 804 pollution monitoring stations placed across 344 cities in India, as part of the National Air Quality Monitoring Program (NAMP)\cite{cpcb}. This means than on an average there is only 1 pollution monitoring station available per 1.6 million people in the country, the density varying from state to state with the North Eastern states having the least number of stations. In contrast there are about 700 Air Quality Monitoring Areas in the UK, which translates to about 1 site per 100,000 individuals \cite{uk aqma}. Even though the NAMP has been running since 1984, the data from the sensors is publicly available only from 2016. The number of stations covering rural areas are very sparse and even the data collected through the monitoring stations is patchy and prone to errors as some of them are manually collected and uploaded \cite{india monitoring}. 

There have been attempts to perform spatial interpolation of ground-based sensors to fill the gaps in the areas where pollution monitoring sites are not present. Various techniques have been employed to create these pollution maps such as the Kriging algorithm, Inverse Distance Weighting\cite{kriging} and more recently Artificial Neural Networks (ANNs)\cite{ann} and Long Short-Term Memory (LSTM) neural networks\cite{lstm} have been used to increase the accuracy of these interpolations. But there are various drawbacks of these models, Inverse Distance Weighting cannot estimate values which fall outside the range of the training data, neural networks do not consider the spatio-temporal associations and the LSTM requires a large set of tagged historical data and plenty of time for the training. Thus the lack of finer resolution data points makes it hard to use spatial interpolation techniques to monitor air quality effectively in India.

The alternative approach is to use remote sensing data to monitor the air quality over a region. The satellites measure the particulate matter present in the atmosphere by using spectroscopic retrieval methods. A significant positive correlation of 0.96 was observed between the Aerosol Optical Thickness (AOT) measured by the MODIS satellite and the $PM2.5$ measurements from the ground\cite{modis aot}. It was possible to get a wider coverage of pollution measure as compared to ground based sensors but this came at a lower spatial resolution, which meant that the data was not adequate for pinpointing the source of the pollution. Spatial scaling techniques were employed to enhance the resolution of the AOT product from 10km to 1km \cite{modis scale} which helped gain a finer insight. The temporal AOT data from Moderate Resolution Imaging Spectroradiometer (MODIS) has been used to perform back trajectory analysis by which the transportation of particulate matter across borders can be traced and the source of the pollutant can be determined \cite{engel}. The Global Ozone Monitoring Experiment (GOME)\cite{gome} was launched in 1995 and it was able to measure tropospheric Ozone and $NO_2$ on a global scale for the first time, but it had a very poor spatial resolution of  80 x 40 km$^2$ per pixel. Apart from this, MODIS has a 1-2 day temporal resolution and GOME has a monthly temporal which meant that data is not available as frequently as the 4hr or 8hr intervals as provided by ground based sensors.

The launch of Sentinel-5P by the European Space Agency (ESA) in October 2017 brought with it an increase in the spectral radius as well as spatial resolution. The on-board TROPOMI sensor can measure $O_{3}$,$NO_{2}$,$SO_{2}$,$CH_{4}$,$CO$,$HCHO$ and $AER_AI$ at a spatial resolution of ${[7x7 km^2]}$, which is about 6 times better as compared to GOME and also improve the sensitivity by an order of magnitude\cite{1}. Sentinel-5P also has a temporal resolution of 1, which meant that it would cover the entire surface area of the earth once every day. Thus, there is now a wider spectrum of pollutants being measured at a very good spatial and temporal resolution, though not as fine granularity as a ground based sensor. Since Sentinel-5P has a wider spectral range and a finer spatial resolution as compared to previous satellites, it helps study the air quality over an area in much finer details and is the preferred choice of data source for experiments carried out in this paper.

Studies have been conducted to analyse the $NO_2$ pollutant, in particular the one conducted by Kaplan et al.\cite{turkeyno2} which correlates it with statistical indicators such as population density. In this study, multiple   pollutants gathered from remote sensing data have been taken into consideration as well and used to perform a clustering of Indian states and districts based on their pollution signatures.

\section{Data Collection} \label{collection}
This section will provide a brief description of how the remote sensing data was obtained. The images were retrieved using Google Earth Engine's Level 3 products for Sentinel-5P. The method by which the Level 2 products, released by the ESA\cite{sentinel}, are processed by Earth Engine are described in the next section.

A yearly average of Level 3 $NO_{2}, SO_{2}$, $CO$, $AER\_AI$, $O_{3}$ and $HCHO$ products was taken over the latitudinal and longitudinal extents of India from January 2019 to December 2019. and was then processed  for further analysis.  The following subsections describe the various Level-2 products that are released by the ESA and explain some of their retrieval methods and their significance in terms of their contribution to air pollution.

\subsection{Nitrogen Dioxide ($NO_2$) }

Sentinel-5P have two sub products which are measured for $NO_2$, namely the tropospheric and the total column. In this study the tropospheric $NO_2$ column, which is the $NO_{2}$ between the surface of the earth and the troposphere, is used as it plays a major role in determining the level of photochemical Ozone \cite{temis}. It must be noted that the TROPOMI  $NO_{2}$ underestimates the  $NO_{2}$ level as compared to ground based sensors but the correlation coefficient was found to be 0.84 and appropriately calibrated, which makes the product accurate enough to be used for analysis. The data measures trace gas concentrations in $mol/m^2$\cite{s5p_no2}.

\subsection{Sulphur Dioxide ($SO_2$) }

The sources of Sulphur Dioxide pollution in the atmosphere are both natural and man-made. The majority of pollution (70\%) arise from coal power plants, smelting industries and mines. Apart from incurring both long term and short term effects on climate, it affects vegetation and water quality when it washes down as acid rain\cite{so2} \cite{s5p_so2}.

\subsection{Aerosol UV Index ($AER\_AI$) }

This product is calculated based on the spectral contrast in the ultraviolet spectral range for the \SI{354}{\milli\metre} and \SI{388}{\milli\metre} wavelengths. It is a long established air quality index monitor and is useful in tracking the plumes released from dust, biomass burning and volcanic ash \cite{s5p_aer}.

\subsection{Carbon Monoxide ($CO$)} 
Carbon monoxide is an important atmospheric trace gas for the understanding of tropospheric chemistry and in certain urban areas, it is a major atmospheric pollutant. In the \SI{2.3}{\micro\metre} spectral range of the shortwave infrared (SWIR) part of the solar spectrum, TROPOMI clear sky observations provide $CO$ total columns with sensitivity to the tropospheric boundary layer. This data has been validated against TCCON and NDACC ground-based network and the MOPITT satellite with a resulting bias of less than 10\%  \cite{s5p_co}.

\subsection{Formaldehyde ($HCHO$)} 
Formaldehyde is an intermediate gas in almost all oxidation chains of non-methane volatile organic compounds (NMVOC), leading eventually to $CO_{2}$. $HCHO$ satellite observations are used in combination with tropospheric chemistry transport models to constrain NMVOC emission inventories in so-called top-down inversion approaches. This data has a mean bias of 50\% with ground based sensors and other satellites such as GOME-2 and OMI and is measured in $mol/cm^{2}$ \cite{s5p_hcho}.

\subsection{Ozone ($O_3$) }
Ozone in the tropical troposphere plays various important roles. The intense UV radiation and high humidity in the tropics stimulate the formation of the hydroxyl radical ($OH$) by the photolysis of $O_3$. The $O_3$ Tropospheric Column gives the measurement of tropospheric ozone between the surface and the 270 hPa pressure level. It is based on the convective cloud differential (ccd)  \cite{s5p_o3}.

\section{Data Preprocessing} \label{preprocessing}
The Earth Engine uses the Level 2 product, from the ESA, and filters the pixels based on minimum pixel quality level corresponding to each scene. These images are then broken into tiles according to the orbit number to make it easier for ingestion and retrieval.

\subsection{Quality Assurance Filtering} 
The quality of the individual observations depends on many factors, including cloud cover, surface albedo, presence of snow-ice, saturation, geometry etc. These observations are filtered in order to avoid misinterpretation of the data quality and to avoid the effects of sun glint. Each of the satellite products comes with a qa\_value (quality assurance) band which can be used to filter out less accurate values. Different thresholds of qa\_value are chosen for different products as defined in the Sentinel-5P Product User Manual.

For making the Level-3 products, Google Earth Engine filters pixels associated with a qa\_value below 0.8 for the $AER\_AI$ product, 0.75 for Tropospheric $NO_2$ and 0.5 for all other products, except $O_3$ for which quality filtering is not done. This takes care of erroneous scenes and problematic retrievals.

\begin{figure}[h]
\centering
\begin{subfigure}{.5\textwidth}
\includegraphics[width=1\linewidth]{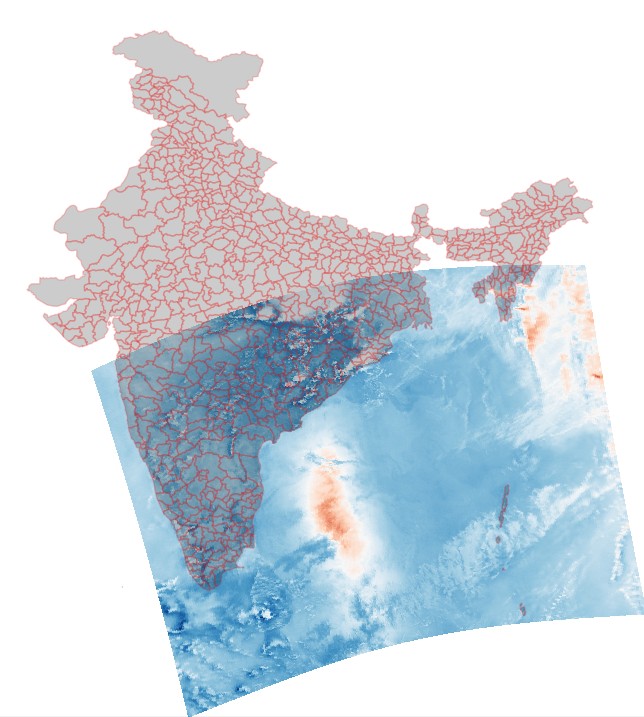}
\caption{Before QA Filtering}
\label{fig:level-2}
\end{subfigure}%
\begin{subfigure}{.5\textwidth}
\includegraphics[width=1\linewidth]{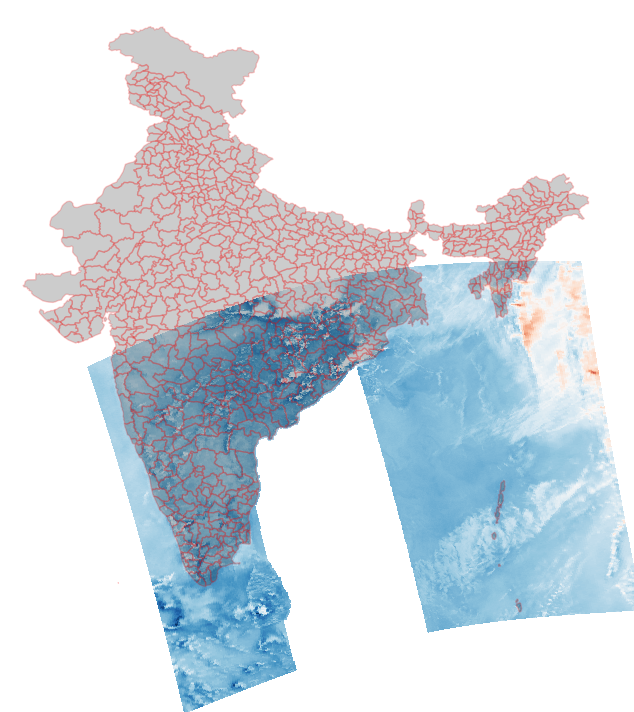}
\caption{After QA Filtering}
\label{fig:level-3}
\end{subfigure}%
\caption{Level-2 $AER\_AI$ Index}
\end{figure}

Fig.\ref{fig:level-2} shows a Level-2 scene of Aerosol Index for a single day released by the ESA. In Fig.\ref{fig:level-3} the missing pixels can be observed after filtering, which correspond to those that had a qa\_value of less than 0.8. 

A yearly average of these Level-3 products was taken in which these missing pixels were filled in with values from those scenes which had data over that region.

\subsection{Regional Masking}
The district level and state level administrative boundaries for India were used to mask the Level-3 products. The $CH_4$ column contained a lot of missing data as the retrieval of this pollutant was of low quality, therefore this pollutant  was dropped in the subsequent analysis. The average pixel values of $NO_2$, $SO_2$, $CO$, $AER\_AI$, $O_3$ and $HCHO$ was calculated for each of the masked district and was stored in a tabular format. A total of 594 districts with six pollutant values was used to frame the monthly pollution data set. This was further cleansed and the rows containing null values were removed.

\begin{figure}[h]
\centering
\includegraphics[width=0.7\textwidth]{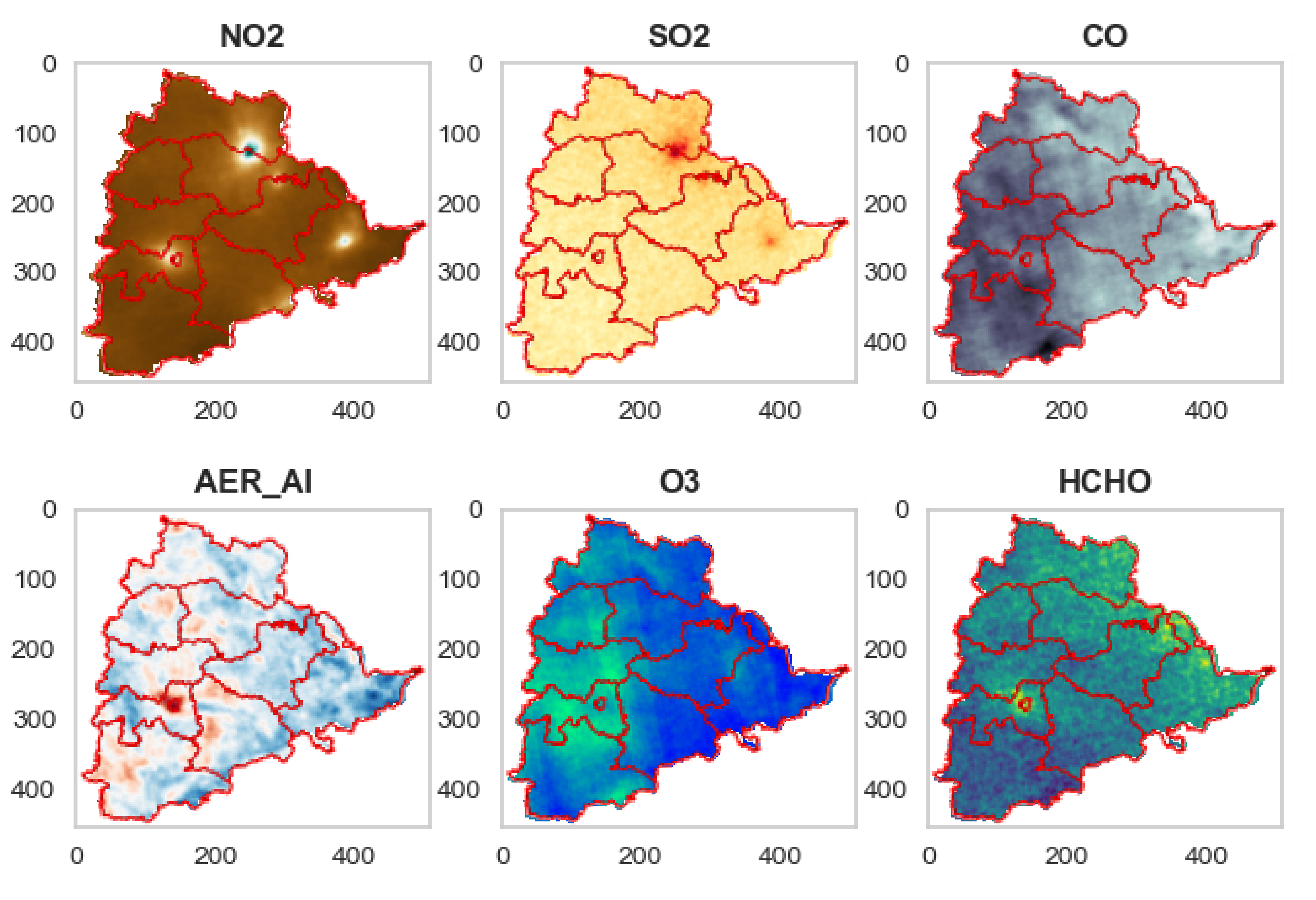}
\caption{Single Level-3 scene masked with Telangana district boundaries}
\end{figure}
Similarly, the Sentinel-5P data was also masked using state border shape files to extract the state-wise average value of pollutants and converted into a tabular data set. The state-wise data had 33 rows and 6 features.
\begin{table}[h]
\begin{adjustwidth}{-.5in}{-.5in}  
        \begin{center}
\begin{tabular}{|l|l|l|l|l|l|l|}
\hline
District        & $NO_2$      & $SO_2$       & CO       & AER\_AI  & $O_3$       & HCHO     \\ \hline
Andaman Islands & 4.28E-05 & -4.73E-08 & 0.037036 & -1.27958 & 0.117128 & 9.89E-05 \\ \hline
Nicobar Islands & 3.89E-05 & -9.24E-06 & 0.033289 & -1.23807 & 0.117677 & 8.95E-05 \\ \hline
Anantapur       & 5.97E-05 & 4.64E-05  & 0.035942 & -1.11059 & 0.116844 & 0.000145 \\ \hline
Chittoor        & 5.81E-05 & 4.95E-05  & 0.035912 & -1.17883 & 0.117122 & 0.000152 \\ \hline
Cuddapah        & 6.20E-05 & 6.00E-05  & 0.037118 & -1.14565 & 0.117214 & 0.000153 \\ \hline
East Godavari   & 6.15E-05 & 8.09E-05  & 0.039594 & -1.17801 & 0.11709  & 0.000174 \\ \hline
Guntur          & 6.56E-05 & 8.16E-05  & 0.039045 & -1.11898 & 0.117253 & 0.000156 \\ \hline
\end{tabular}
    \end{center}
\end{adjustwidth}
\caption{A Snippet of the District-wise Pollution Data}
\end{table}
\subsection{Standardization}
The dataset was then re-scaled by first removing the mean of each pollutant and scaling each column by the standard deviation as shown in Eqn.\ref{eqn:standardization}. This ensures that the features are of comparable denominations before running the clustering algorithms on them.
 \begin{equation}
     z = \frac{x-\overline{x}}{\sigma}
     \label{eqn:standardization}
 \end{equation}

\section{Clustering}\label{clustering}
Clustering is a method of grouping based on patterns in the data \cite{clustering}. This technique is primarily used to find clusters of data points with inherent similarity in unlabelled datasets. In this case, the unlabelled dataset consists of different pollutants emitted ($NO_2$, $SO_2$, $CO$, $AER\_AI$, $O_3$ and HCHO) for each state or district in India. The aim is to find clusters comprising of states or districts that emit a similar pollution signature which will help isolate pollution sources more easily.

\subsection{Clustering Methods}
Unsupervised clustering was performed on the pollution datasets using three different algorithms, namely - K-Means clustering, Agglomerative clustering and DBSCAN \cite{clustering techniques}. The distance measure which was used in all three methods was the L-2 norm, Euclidean Distance.

\subsubsection{K-Means Clustering}
Partitioning based unsupervised clustering method reallocate data points by moving them from one cluster to another, starting from an initial partitioning. This algorithm works by initializing K points as cluster centers. In each iteration, every point in the dataset is assigned to the cluster it is closest to (using the L2 norm distance). The cluster center is then reinitialized to the mean of the cluster set and the clustering iteration until convergence is achieved. 

\subsubsection{Ward Agglomerative Clustering}
Hierarchical clustering constructs the clusters by recursively splitting or combining the data points. In agglomerative clustering (a method within the broader class of hierarchical clustering methods)\cite{wards}, the clusters are built by iteratively merging smaller clusters starting from individual data point up until the required number of clusters are reached. The Ward's distance minimizes the overall inter-cluster sum of squared distances within all clusters.
\begin{equation}
     \Delta \big(A, B\big) = \sum_{i\in {A\cup B}} \|\vec{x}_i-\vec{m}_{A\cup B}\|^2 - \sum_{i\in A} \Vert\vec{x}_i-\vec{m}_{A}\Vert^2 - \sum_{i\in B} \Vert\vec{x}_i-\vec{m}_{B}\Vert^2
 \end{equation}
Here, $\vec{x}_i$ represents the center of the $i^{th}$ cluster and $\Delta \big(A, B\big)$ represents the cost of merging clusters A and B.

\subsubsection{DBSCAN Clustering}
Density based clustering tries to associate each point to a set of probability distributions. This algorithm does not take the number of clusters as an input and uses two parameters $min\_pts$ and $epsilon$ to form clusters. $Epsilon$ determines the maximum distance between two points upto which they can be grouped within the same cluster and the $min\_pts$ determines the minimum number of points that must fall within a cluster for it to not be designated as a noise point. It is useful in detecting noise and outliers in the data.

\subsection{Optimal Number of Clusters} 
The elbow method \cite{cluster no} is a technique used to determine the optimal number of clusters to be chosen based on heuristics such as inter cluster similarity and intra-cluster similarity.  In the method presented here, the number of clusters is iteratively increased from 2 till 15 and the point at which the graph of the cost function has the highest curvature is taken as the elbow, or optimal number of clusters.

The elbow method was performed using the distortion score as the cost function to find the optimal number of clusters for K-means clustering and the silhouette score was used to ensure that the intra-cluster similarity was optimal. Both scores are explained below. The same number of clusters as derived from the elbow method was used to slice the hierarchical clustering to compare the results of the two algorithms.

\subsubsection{Distortion Score}
 This metric give information about the overall cluster dissimilarity. It is calculated as the mean sum of squared distances to centers.
 \begin{equation}
     S = \sum_{i=1}^{n} (x_i-\overline{x})
     \label{eqn:dist}
 \end{equation}
In Eqn.\ref{eqn:dist}, x\textsubscript{i} represents the i\textsuperscript{th} row of the dataset and $\bar{x}$ represents the mean of the cluster it belongs to. the lower the value of S, lower the dissimilarity, the more optimal is the solution.
\begin{figure}[h]
    \centering
    \includegraphics[width=0.6\textwidth]{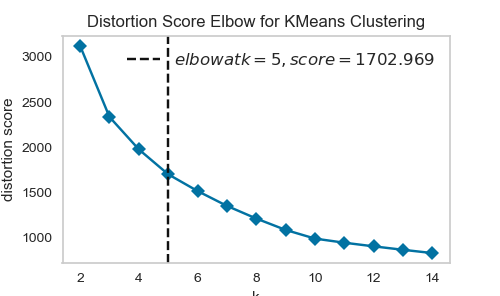}
    \caption{Distortion Score Elbow for K-Means}
    \label{fig:dist}
\end{figure}
Fig.\ref{fig:dist} represents the plot for the distortion scores as a result of fitting the K-Means model to the dataset, while varying the value of K from 2-15. The elbow method analysis results in the elbow line being drawn at K=5 as this is the point of maximum curvature in the curve.

\subsubsection{Silhouette Score}
The silhouette score metric represents the intra-cluster similarity.
It is calculated as the mean ratio of intra-cluster and next nearest-cluster distance.

\begin{equation}
     S = \frac{b - a}{max(a,b)}
     \label{eqn:silh}
\end{equation}
In Eqn.\ref{eqn:silh}, ${a}$ is the mean distance between a sample and all other points in the same class and $b$ is the mean distance between a sample and all other points in the next nearest cluster. The score is higher when clusters are dense and well separated.
\begin{figure}[h]
    \centering
    \includegraphics[width=0.6\textwidth]{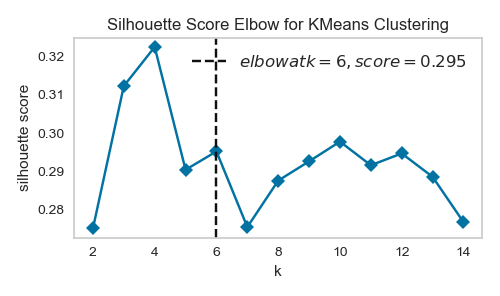}
    \caption{Silhouette score Elbow  for K-Means}
    \label{fig:silh}
\end{figure}

Fig.\ref{fig:silh} represents the plot for silhouette scores as a result of fitting the K-Means model to the dataset while varying the value of K from 2-15. The elbow method analysis results in the elbow line being drawn at K=6.

The elbow method using distortion score resulted in 5 being the optimal number of clusters. From the silhoutte score plot, it can be inferred that the intra-cluster similarity of K=5 was 0.290 which was not too far away from the optimal silhouette score of 0.295. Hence, K=5 was chosen as the optimal number of clusters to prevent over-fitting on the data and get well rounded clusters.

\subsection{Cluster Validation}
The silhouette method was used to determine the intra-cluster similarity of the clusters formed by the three algorithms mentioned in the previous section. It gives an idea of how closely related a state or district is to the cluster it has been assigned to by the respective method. Values close to 1 indicate a high affinity to that cluster and negative values imply that the data point might have been wrongly assigned to that cluster.

\begin{figure}[h]
    \centering
    \includegraphics[width=0.7\textwidth]{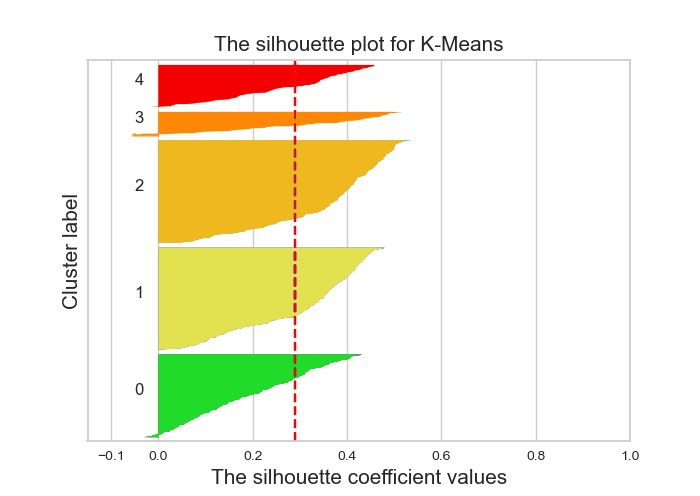}
    \caption{Silhouette Score for K-Means}
    \label{fig:silhp}
\end{figure}

Each color in Fig.\ref{fig:silhp} represents the distribution of the silhouette score of each state that falls within the cluster. The wider the graph for a cluster, the more the number of states that fall into it. The average silhouette score for K-Means was 0.290 and for Ward Agglomerative was 0.280. 

The lower polluting states and the higher polluting states had silhouette coefficients of almost 0.5, which indicates a higher intra-cluster similarity. Both K-Means and Ward Agglomerative clustering resulted in a similar distribution of states across clusters. Only the state of Telangana was placed in cluster 1 in K-Means and cluster 0 in Ward Agglomerative. In the case of DBSCAN, it does not need to specify the number of clusters and instead the values of the parameter $min\_pts$ was fixed to 3 and that of $epsilon$ was 1.7. DBSCAN Clustering resulted in a few states being classified as outliers since their pollution signatures did not match with any other state. But the remaining clusters that were formed were in correlation with K-Means and Ward Agglomerative clustering. The results from K-Means clustering have been presented and used for analysis in the following section.

\subsection{Analysis of Pollution Signatures and Clustering Results across Indian States and District}
In this section, the various results from state as well as district wise clustering and the corresponding pollution signatures obtained for different clusters have been presented and analysed in depth.

The bar plots shown below in Fig.\ref{fig:signature} represent the average pollution signatures for every cluster obtained as a clustering of pollution data across different states and districts. These pollution signatures serve as a unique representation for each cluster. Each pollution signature shows the average pollutant magnitude for a given cluster. States and Districts which are part of Cluster 0, on average, have the lowest pollution profile and those which fall into Cluster 4 emit the highest level of pollutants as can be seen in Fig.\ref{fig:signature}. The varying  trends for each pollutant across each cluster for both state-wise and district-wise clustering have also been shown in Fig.\ref{fig:trends}. Fig.\ref{fig:state} shows the state-wise cluster map and Fig.\ref{fig:district} shows the district-wise cluster map across India as obtained from K-means clustering algorithm. 

\begin{figure}[h]
    \centering
    \begin{subfigure}{.5\textwidth}
        \includegraphics[width=0.95\linewidth]{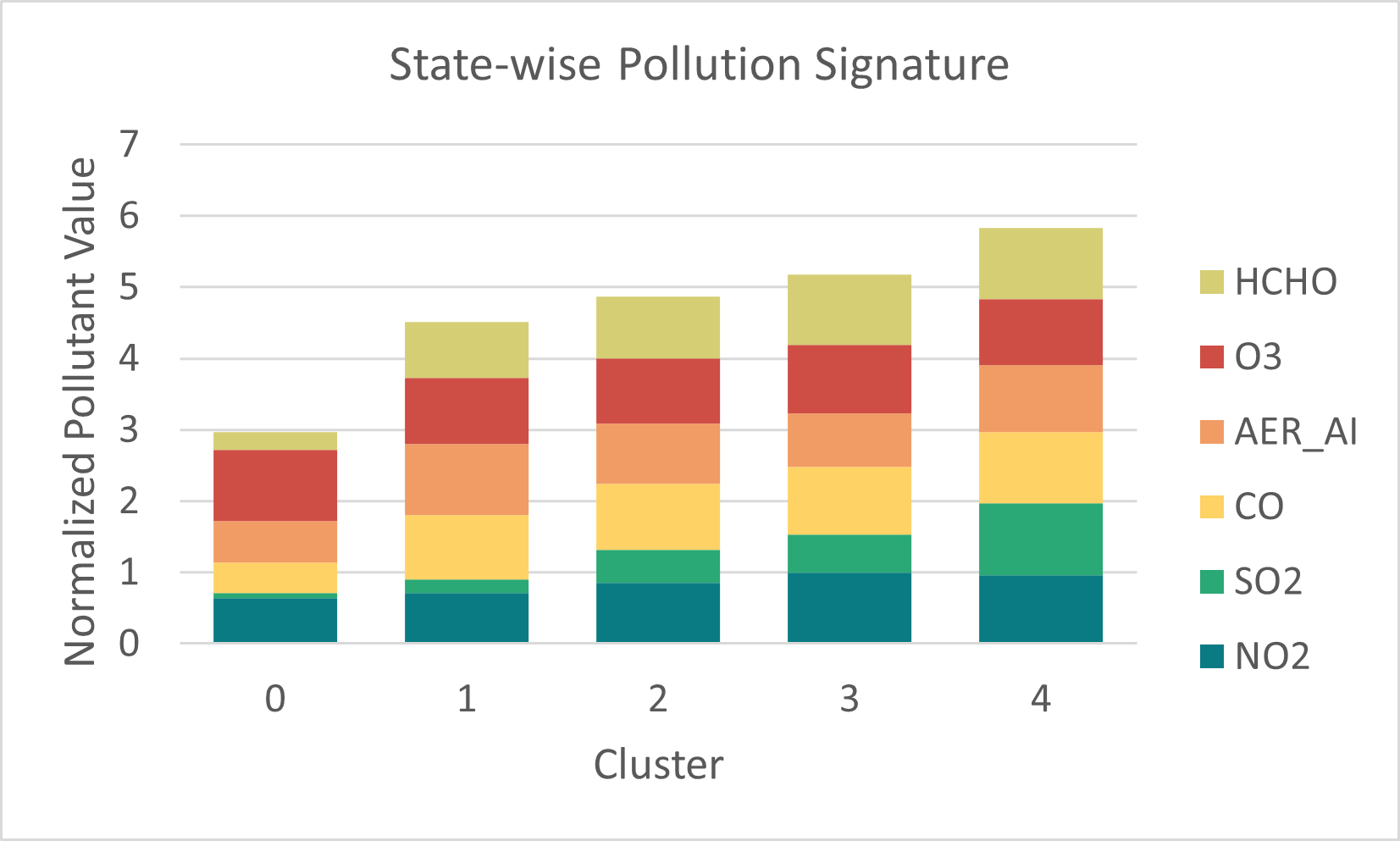}
        \caption{State-wise Pollution Signatures}
    \end{subfigure}%
    \begin{subfigure}{.5\textwidth}
        \includegraphics[width=0.95\linewidth]{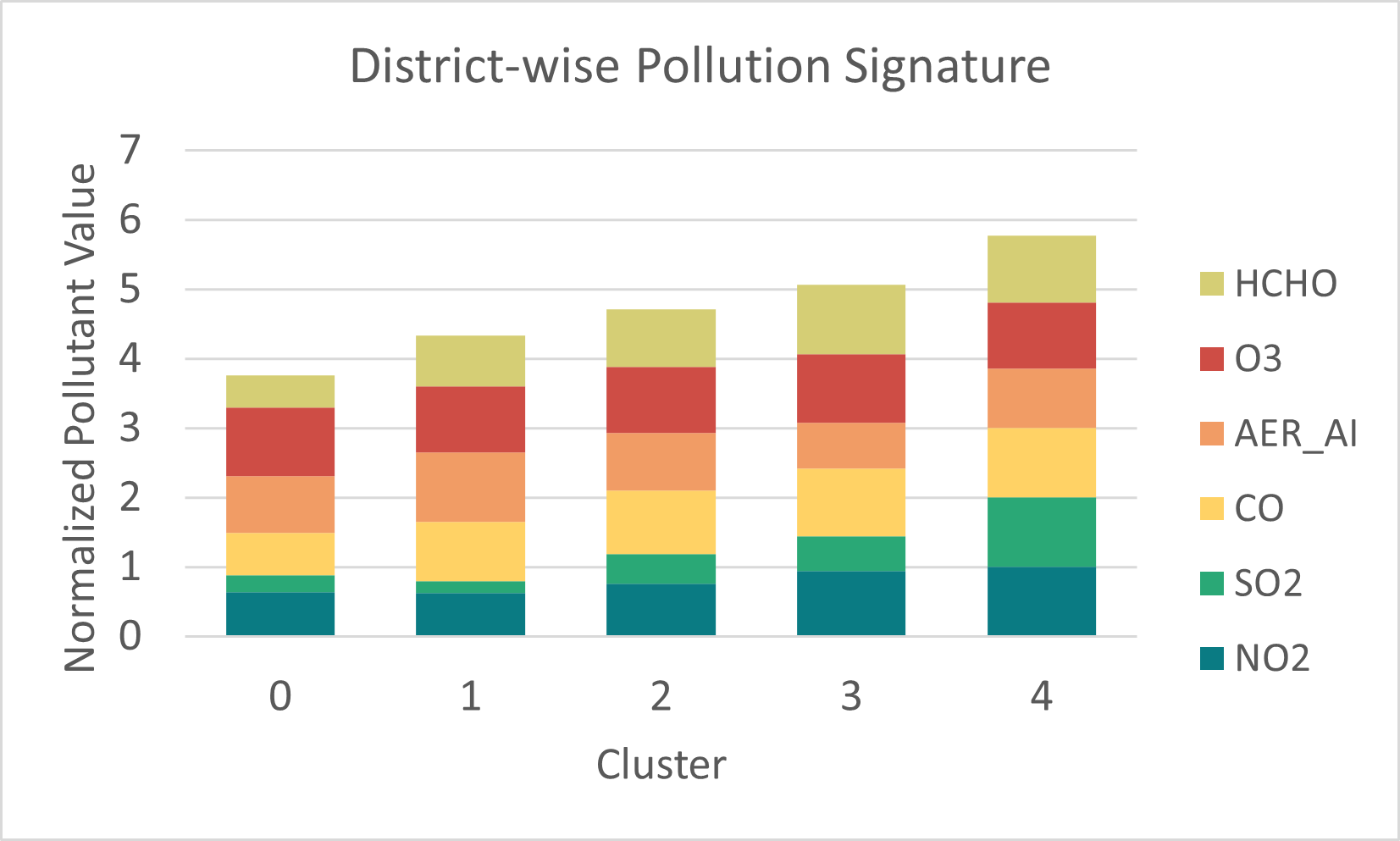}
        \caption{District-wise Pollution Signatures}
    \end{subfigure}%
    \caption{Pollution Signatures for K-Means Clustering}
    \label{fig:signature}
\end{figure}%

\begin{figure}[h]
    \centering
    \begin{subfigure}{.5\textwidth}
        \includegraphics[width=0.95\linewidth]{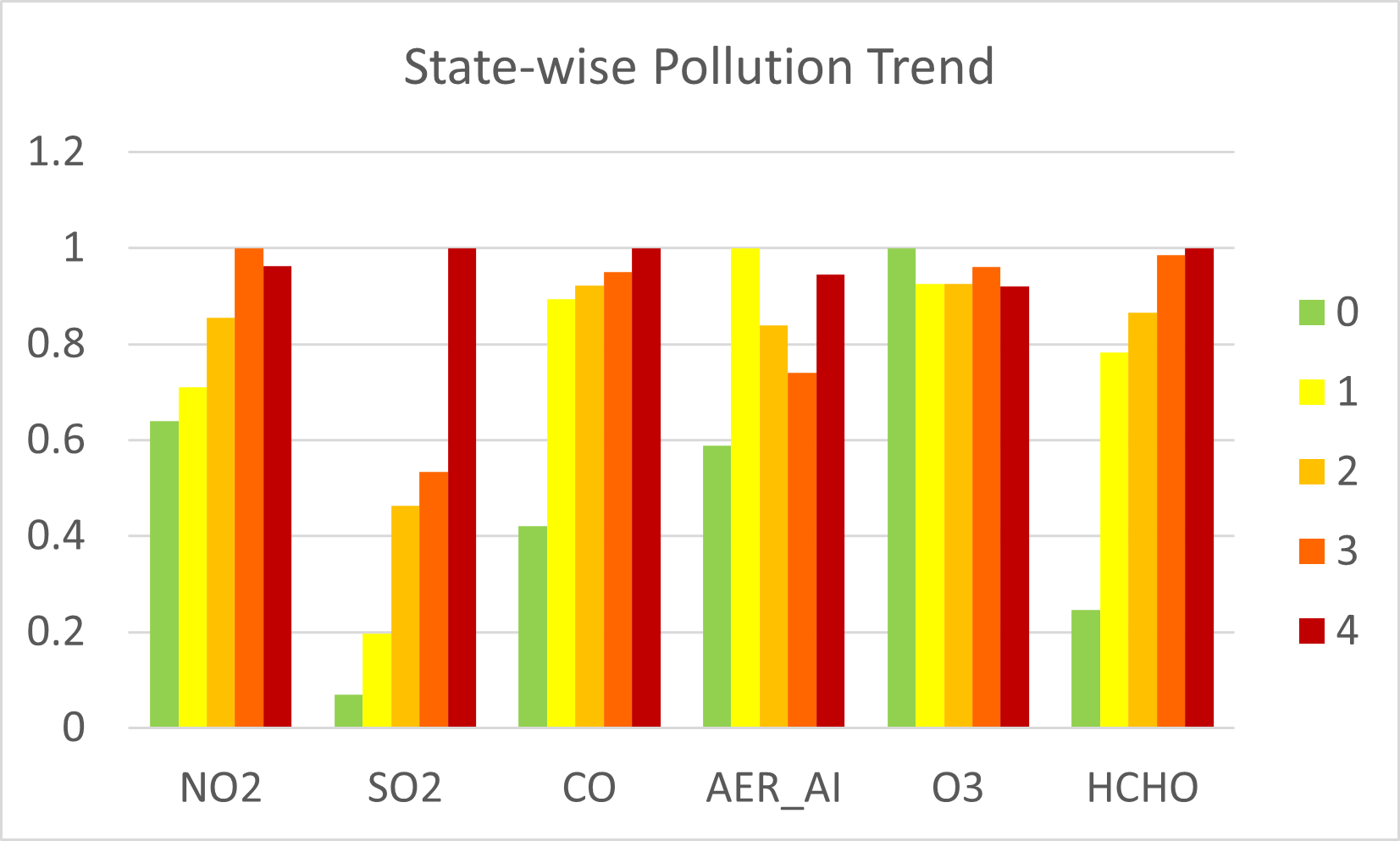}
        \caption{State-wise Pollution Trend}
    \end{subfigure}%
    \begin{subfigure}{.5\textwidth}
        \includegraphics[width=0.95\linewidth]{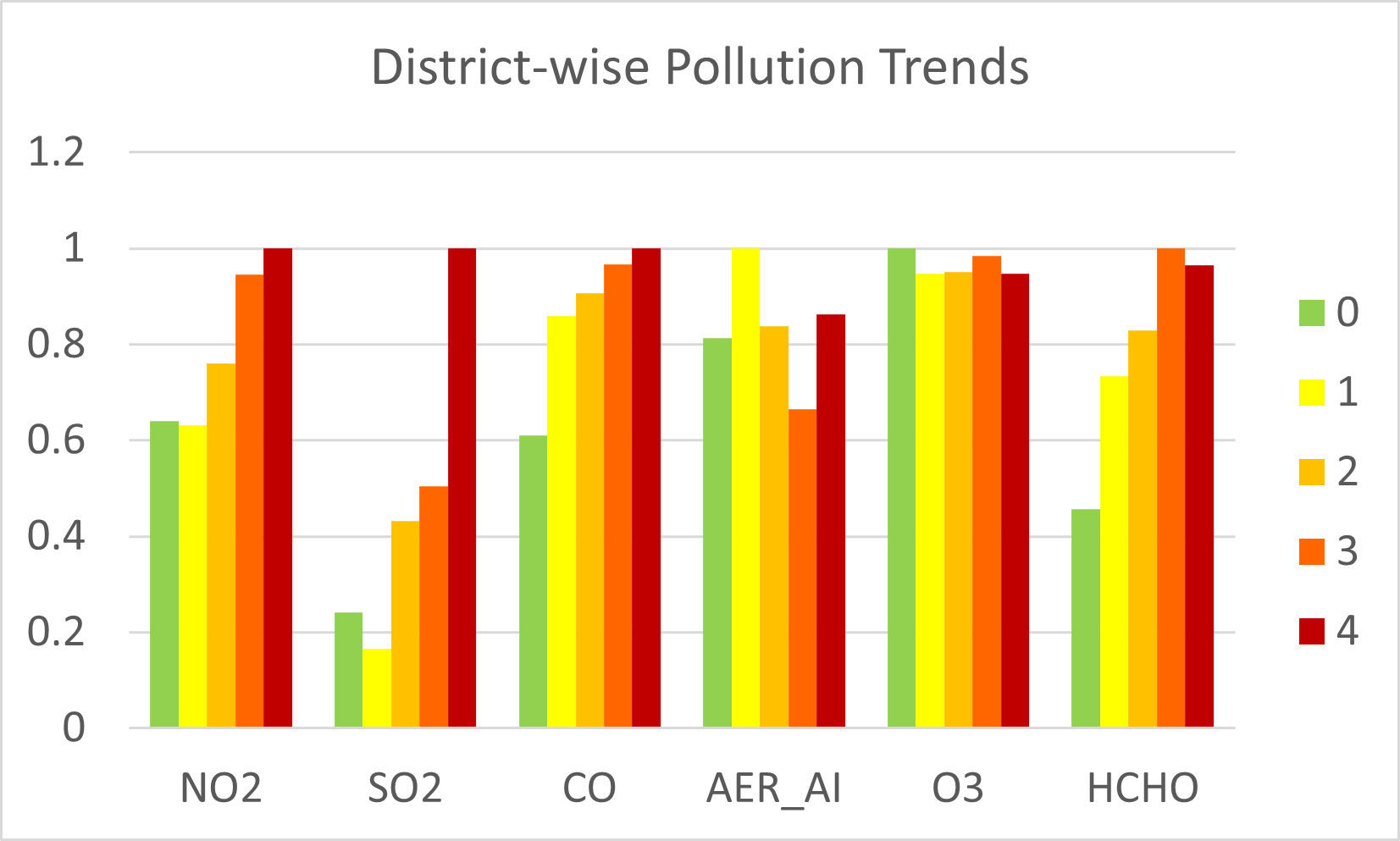}
        \caption{District-wise Pollution Trend}
    \end{subfigure}%
    \caption{Varying Pollution Trends across K-Means Clusters}
    \label{fig:trends}
\end{figure}%

Each of the cluster behaviour can be explained in terms of its corresponding pollution signature and trends as follows:

Cluster 0 - As can be seen in Fig.\ref{fig:signature}, cluster 0 has the least concentration of pollutants.
States such as Jammu \& Kashmir, Uttarkhand, Himachal Pradesh and Sikkim fall under this cluster. These states have a very low population density and consequently low vehicular traffic as they mainly comprise of mountainous terrains. These states also have significantly lower industrial activity and hence do not contribute much to the air pollution.

Cluster 1 - On an average, the regions which fall under this cluster have a higher level of emission of all pollutants when compared to cluster 0, except for $SO_2$ as can be seen in Fig.\ref{fig:trends}.
This cluster comprises of the southern states and a majority of the north eastern states.

Cluster 2 -The western and central states such as Telangana, Odisha, Gujarat, Madhya Pradesh and Maharashtra fall under this cluster. It can be noted that in the district-wise clustering in Fig.8b, the city of Chennai falls under this cluster 2, whereas it came under cluster 1 along with the state of Tamil Nadu in Fig.\ref{fig:state}.

Cluster 3 - This is the second highest polluting cluster and most of the states which fall under this have a lot of industrial presence which is attributed with high pollutant and in particular high $NO_2$ emissions.  States such as Uttar Pradesh, Haryana, Rajasthan and Punjab fall under this in the state-wise clustering.

Cluster 4 - This is the cluster which has the higher average percentage of pollutants $SO_2$ and $NO_2$ as can be seen in Fig.\ref{fig:trends}. These states have some of the highest population densities in India and contain some of the worst polluting cities in the world. It is worth noting here that the results are consistent with the report from TERI - The Energy and Resources Institute \cite{teri} , where the states belonging to the cluster 3 and cluster 4 are those which have the highest $PM10$ emissions from brick kilns as well as from coal and iron ore mining.

The biggest difference in terms of state vs. district level clustering can be noted from the regions which fall under this cluster . In Fig.\ref{fig:state} it can be seen that only the state Delhi falls under this cluster 4, but in the district-wise clustering as shown in Fig.\ref{fig:district}, it can be seen that a majority of the districts from Uttar Pradesh, Bihar, Haryana, Punjab and Rajasthan as well now come under cluster 4, representing the highest overall pollution signature. 

\begin{figure}[h]
    \centering
    \begin{subfigure}{.5\textwidth}
        \includegraphics[width=0.95\linewidth]{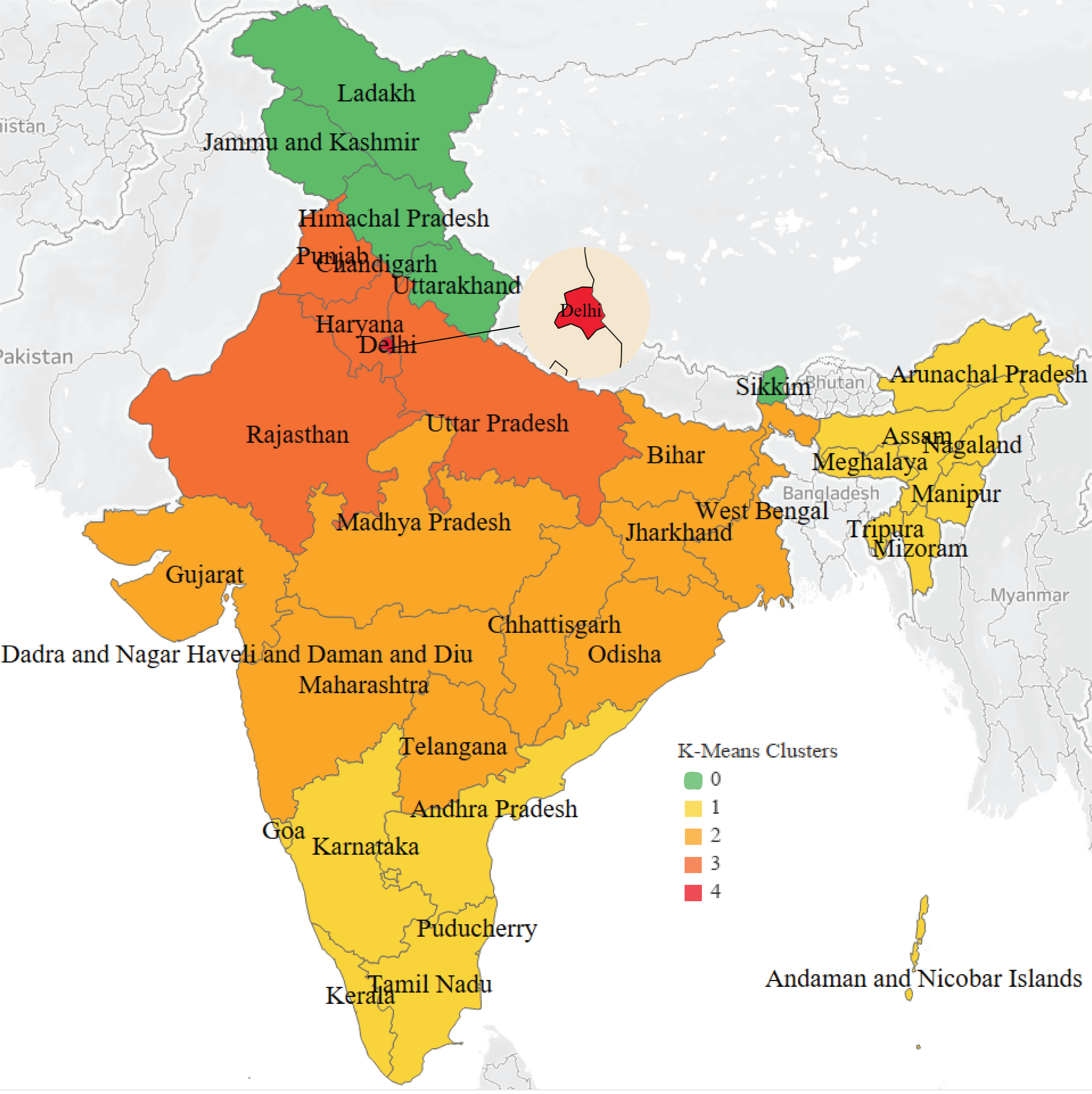}
        \caption{State-wise Clusters}
        \label{fig:state}
    \end{subfigure}%
    \begin{subfigure}{.5\textwidth}
        \includegraphics[width=0.95\linewidth]{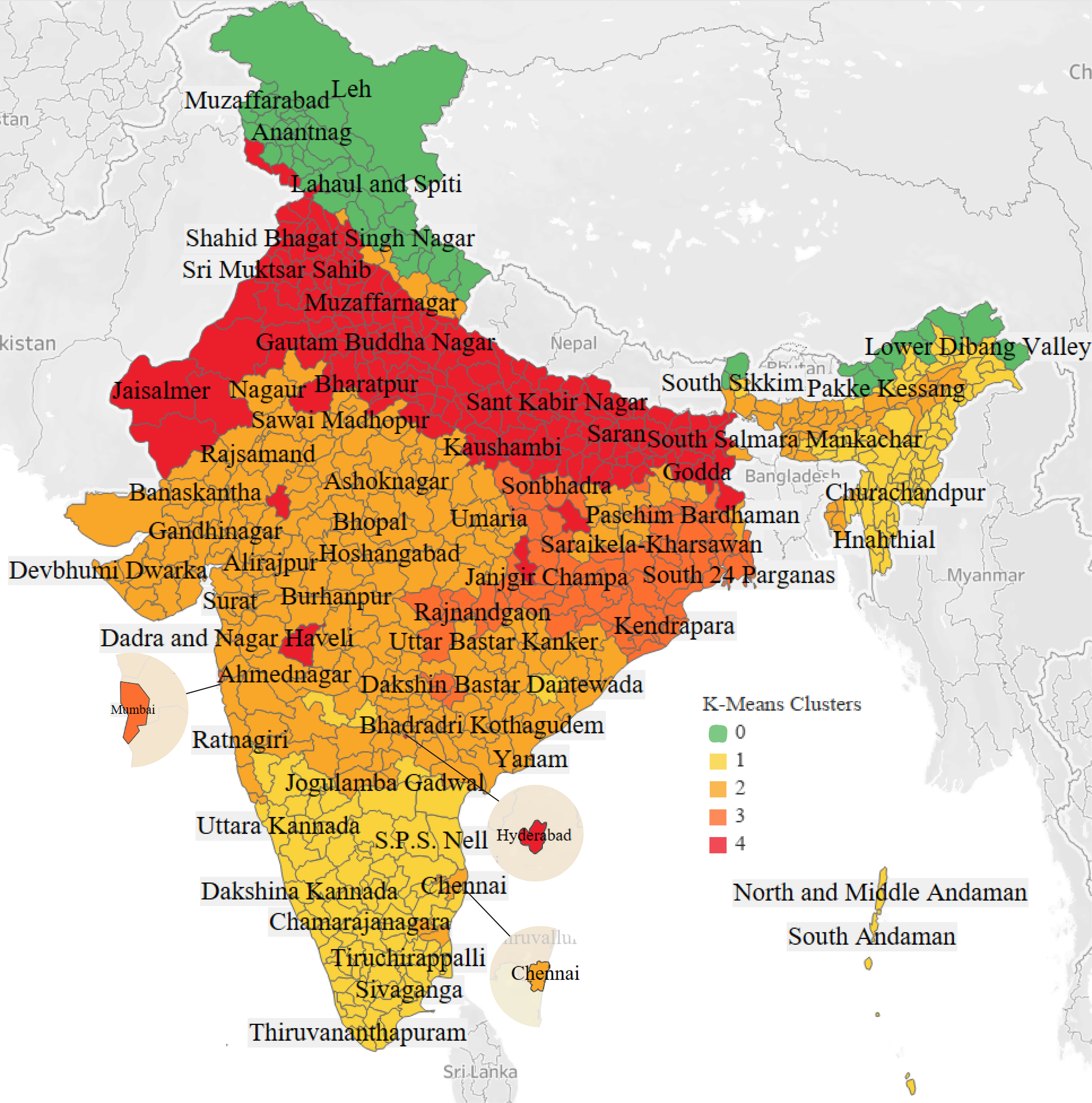}
        \caption{District-wise Clusters}
        \label{fig:district}
    \end{subfigure}%
    \caption{Visualization of K-Means Clusters}
\end{figure}%

Further comparing the two maps in Fig.\ref{fig:state} it can be seen that the mining intensive states of Jharkhand, West Bengal, parts of Chhattisgarh and Odisha fall under cluster 2, whereas in Fig.\ref{fig:district} the districts from these states fall under cluster 3, which indicates that they belong to a lower polluting category. Upon comparing the state-wise and district-wise clustering, we can see that cities like Mumbai, Chennai and Hyderabad fall under a higher polluting cluster as compared to other districts within the same state. This is in line with expectations since these urban cities tend to emit higher levels of pollution. Thus based on this analysis, it is observed that a finer resolution regional pollution classification can be seen in district-wise clustering as compared to the state-wise clustering.

\section{Conclusions and Future Work} \label{conclusion}
Acknowledging the importance of managing pollution levels so that it does not affect the socio-economic and health status of the general population, it is important to understand the precise pollutant signature over a certain area and the possible sources of the pollutants. The presented study explored three clustering algorithms on data retrieved from ESA's Sentinel-5P satellite to address this issue. The clustering algorithms were used to assign unique pollutant signatures to states and districts across India. The results have shown to be promising. To take this work further, it is planned to improve on the clustering algorithms to understand, if a similar or higher accuracy can be attained at even finer granularity than a district level. In addition, studies need to be conducted that will help correlating pollution levels with socio-economic factors of the region. Furthermore, there is a need to study the affect of other variables such as wind, atmospheric pressure etc. to understand the transport of pollutants from its source \cite{pollution trajectory}. This will help further identify pollution sources with higher accuracy.

\end{document}